\documentclass[sigconf, screen]{acmart}
\AtBeginDocument{%
  }

\setcopyright{acmlicensed}
\copyrightyear{2026}
\acmYear{2026}
\acmDOI{XXXXXXX.XXXXXXX}
\acmConference[Conference acronym 'XX]{Make sure to enter the correct
  conference title from your rights confirmation email}{June 03--05,
  2018}{Woodstock, NY}
\acmISBN{978-1-4503-XXXX-X/2018/06}

\acmSubmissionID{1758} 

\renewcommand\footnotetextcopyrightpermission[1]{}
\setcopyright{none}

\usepackage{multirow}
\usepackage{colortbl}
\usepackage{tabularx}
\usepackage{amsmath, bm}
\usepackage{enumitem}
\newcommand{\ptitle}[1]{%
  \par\smallskip
  \noindent\textbf{#1}\hspace{0.5em}%
}

\usepackage{tcolorbox}
\tcbuselibrary{skins, breakable, theorems}

\definecolor{linecolor1}{RGB}{246, 248, 239}
\definecolor{linecolor2}{RGB}{230, 234, 217}
\definecolor{linecolor3}{RGB}{211, 222, 190}
\newcolumntype{x}[1]{>{\centering\arraybackslash}p{#1pt}}

\begin{document}

\title{VeriSpace: Spatially Grounded Action Verification for Vision-Language-Action Models}

\author{Guiyu Zhao$^{1,2}$, Longteng Guo$^{1}$, Junyou Zhu$^{1,2}$, Jun Fu$^{1}$, Yanghong Mei$^{1,2}$,  Bin Cao$^{1,2}$, \\ Jie Jiang$^{1}$, Xingjian He$^{1}$, Jing Liu$^{1,2,\dagger}$}

\affiliation{%
  \institution{$^{1}$Institute of Automation, Chinese Academy of Sciences, Beijing, China} 
  \country{} 
}

\affiliation{%
  \institution{$^{2}$University of Chinese Academy of Sciences, Beijing, China} 
  \country{}
}


\begin{abstract}
Vision-language-action (VLA) models have shown strong promise for robotic manipulation, but their reliability at test time remains limited by one-shot action prediction, where even small action errors can cause grasp failure, collision, or incorrect task progression. A natural alternative is to equip VLA systems with test-time verification, allowing multiple candidate actions to be proposed and evaluated before execution. However, reliable action verification is challenging because it requires not only distinguishing subtle geometric differences between candidate actions, but also assessing whether an action makes meaningful progress toward the task goal. We present VeriSpace, a 3D-aware action verifier for test-time action selection in VLA systems. VeriSpace evaluates candidate actions through two key components: Dual-Path 3D-Injected Scene Encoding, which constructs a scene representation that jointly preserves visual semantics and explicit 3D geometry, and Spatially-Grounded Action Reasoning, which evaluates each action by reasoning over task-relevant spatial relations, geometric validity, and expected goal progress. Together, these components enable more reliable discrimination between subtle yet outcome-critical action candidates while remaining fully compatible with existing VLA policies. Experiments on public benchmarks and real-world robotic manipulation tasks show that VeriSpace consistently improves decision reliability over both underlying VLA policies and prior verification-based methods, yielding substantial gains in both in-distribution and out-of-distribution settings.

\end{abstract}





\begin{teaserfigure}
  \centering
  \includegraphics[width=\textwidth]{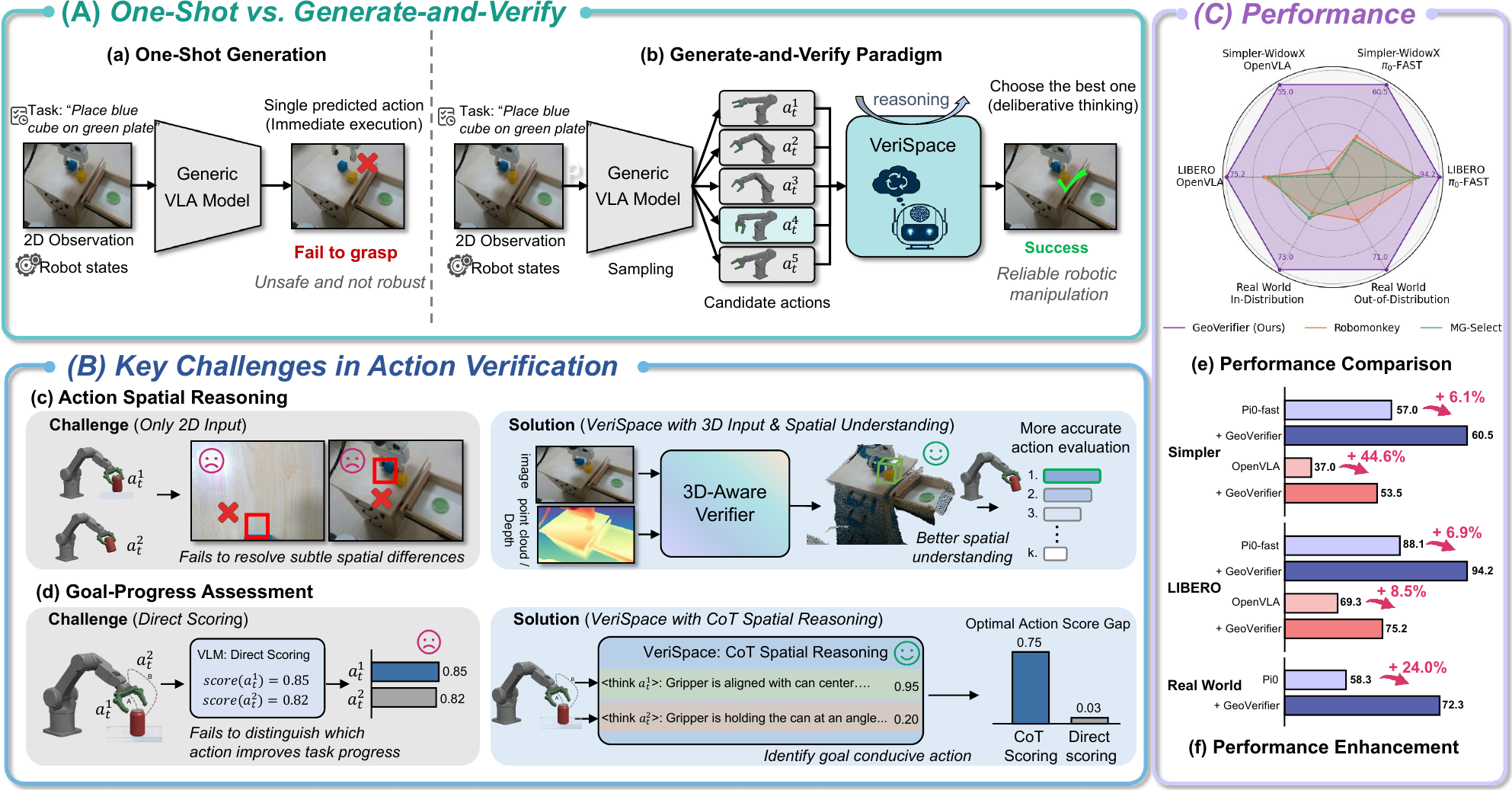}
    \caption{
    \textbf{Towards Reliable Test-Time Action Verification.}
    (A) Instead of directly executing a single predicted action, VeriSpace adopts a test-time \emph{generate-and-verify} paradigm for robust action selection.
    (B) Reliable action verification in robotic manipulation requires both fine-grained spatial reasoning and goal-progress assessment, calling for 3D-aware perception and explicit spatial reasoning.
    (C) VeriSpace delivers consistent gains across public benchmarks, VLA backbones, and real-world tasks.
    }
  \label{fig:teaser}
\end{teaserfigure}


\maketitle

\section{Introduction}
\label{sec:intro}
Vision-language-action (VLA) models~\cite{brohan2022rt, zitkovich2023rt, kim2024openvla, black2024pi_0, intelligence2025pi_, sun2025geovla} have emerged as a promising paradigm for robotic manipulation, enabling robots to translate high-level language instructions and visual observations into executable actions. 
Powered by large-scale robotic datasets~\cite{khazatsky2024droid, walke2023bridgedata, wu2024robomind, liu2023libero} and multimodal foundation models, VLA systems have shown strong performance and encouraging generalization across diverse manipulation tasks.
Despite this progress, reliable deployment of VLA models in real-world environments remains challenging. 
Unlike recognition or language generation tasks, robotic manipulation requires the model to commit to a physical action whose quality directly affects task success, safety, and downstream state evolution.
Even small errors in action prediction can lead to grasp failure, object collision, or incorrect task progression, especially in unstructured scenes and tasks requiring fine-grained spatial control. 
This makes robust test-time decision-making a central requirement for VLA systems.

Yet, most existing VLA systems still operate in a largely one-shot manner at inference time: given the current observation and instruction, the model predicts a single action and immediately commits to execution. 
While this paradigm is efficient, it leaves little room for deliberation when the scene is ambiguous, the manipulation geometry is challenging, or multiple plausible actions exist. 
In such cases, a more capable robotic system should be able to allocate additional computation at test time, consider alternative action hypotheses, and select the one most likely to succeed before interacting with the physical world. 
This motivates a test-time scaling paradigm for VLA models, in which candidate actions are first proposed and then evaluated prior to execution. 
Under this paradigm, the quality of the final decision depends critically on the verifier’s ability to distinguish successful actions from subtle but consequential failures. 

However, making reliable action verification work in practice is far from straightforward. 
In robotic manipulation, the difficulty often lies in two places. 
\textbf{(1) Action Spatial Reasoning:} Candidate actions that look almost the same can lead to very different outcomes once executed. 
A small shift in position, a slight change in orientation, or a narrow difference in clearance or contact can determine whether the robot grasps an object cleanly, collides with the scene, or misses the target entirely. 
As a result, action verification depends on understanding the fine-grained spatial structure of the scene and how each action interacts with task-relevant objects. 
This makes spatially grounded perception essential, and gives 3D information a natural role in evaluating candidate actions. 
\textbf{(2) Goal-Progress Assessment:} A candidate action is also meaningful only in the context of where the task is supposed to go next. 
The same motion may be useful in one stage of a task and unhelpful in another, depending on whether it moves the robot and the scene closer to successful completion. 
Reliable verification therefore depends on judging how much progress an action makes toward the task goal, based on the current scene, the intended target state, and the transition between them. 
These subtle differences are often difficult to capture through direct scoring alone.

To address these challenges, we propose \textbf{VeriSpace}, a \textbf{3D-aware action verifier} for improving test-time action selection in VLA systems through spatially grounded evaluation. Unlike prior approaches that rely on direct action scoring, VeriSpace explicitly models both scene geometry and task-driven spatial reasoning to evaluate candidate actions. Specifically, it is built upon two complementary components: \textbf{Dual-Path 3D-Injected Scene Encoding}, which constructs a structured scene representation that jointly preserves visual semantics and explicit 3D geometry, enabling the verifier to resolve fine-grained spatial differences that are critical for manipulation; and \textbf{Spatially-Grounded Action Reasoning}, which evaluates each candidate action by reasoning over task-relevant spatial relations, assessing both its geometric validity and its expected contribution toward task completion. By combining geometry-aware perception with structured reasoning, VeriSpace transforms action verification from direct scoring into a more reliable decision process, allowing it to discriminate between subtle yet outcome-critical action candidates while remaining fully compatible with existing VLA policies without modifying their training or architecture.

We evaluate VeriSpace on public benchmarks~\cite{li24simpler, liu2023libero} and real-world robotic manipulation tasks. Across all settings, VeriSpace consistently improves the reliability of VLA decision-making and delivers strong gains over both the underlying policy models~\cite{kim2024openvla, black2024pi_0} and prior verification-based methods~\cite{kwok2025robomonkey, jang2025verifier}. On the SIMPLER benchmark, VeriSpace improves OpenVLA~\cite{kim2024openvla} by 18.0 percentage points and surpasses the previous state of the art by 13.0 points. In real-world experiments, it yields substantial improvements in both in-distribution and out-of-distribution settings, demonstrating that the benefits of test-time verification extend beyond simulation. VeriSpace also generalizes well across different VLA backbones, showing consistent gains as a plug-and-play verifier.


In summary, the main contributions of this paper are three-fold:
\begin{itemize}[leftmargin=*]
    \item We introduce VeriSpace, a 3D-Aware Action Verifier that improves test-time action selection for VLA models through more reliable spatial reasoning, while remaining fully plug-and-play with existing VLA policies.
    
    \item We develop Dual-Path 3D-Injected Scene Encoding and Spatially-Grounded Action Reasoning, enabling the verifier to better understand scene geometry and assess how candidate actions move the task toward completion.
    
    \item We validate VeriSpace on public benchmarks and real-world robotic manipulation tasks, showing state-of-the-art performance, broad compatibility across VLA backbones, and strong transfer to real-world settings.
\end{itemize}

\section{Related Work}

\ptitle{Vision–Language–Action Models.}
With the rapid development of large language models~\cite{touvron2023llama, brown2020language, ouyang2022training} and multimodal large language models~\cite{GPT4Vision23,llava23,DreamLLM23,ShapeLLM24}, Vision-Language-Action (VLA) models have gradually become a mainstream approach for general robotic manipulation. Pioneering works like the RT series~\cite{brohan2022rt, zitkovich2023rt, belkhale2024rt} and OpenVLA~\cite{kim2024openvla} have demonstrated strong instruction-following and generalization capabilities by fine-tuning MLLMs on large-scale robotic manipulation datasets. To further enhance continuous control and action precision, recent advancements have integrated powerful generative architectures into the VLA framework; for example, RDT~\cite{liu2024rdt} incorporates a scalable Transformer and a unified action space within a diffusion formulation to handle heterogeneous multi-modal inputs across diverse robot embodiments, while $\pi_0$~\cite{black2024pi_0} augments pre-trained VLMs~\cite{beyer2024paligemma} with continuous action outputs via flow matching, enabling high-frequency and dexterous policy generation.
Based on these works~\cite{brohan2022rt, black2024pi_0, kim2024openvla}, many advanced methods~\cite{wen25tinyvla,kim2025fine,qu2025spatialvla,robovlms24,liu2025hybridvla,intelligence2025pi_,li2026pointvla,zhang20254d,li2025bridgevla,zhang2025spatial,sun2025geovla,lin2025evo} have been developed to achieve improved performance.
Although these methods~\cite{zhang20254d,intelligence2025pi_,qu2025spatialvla,li2025bridgevla,zhang2025spatial,wen25tinyvla} achieve promising results, the absence of test-time validation~ limits their ability to ensure robust robot manipulation, particularly in real-world and unseen scenarios.



\ptitle{Test-Time Verification for Robotics.}
Introducing additional computation at test time has achieved strong performance gains across various fields of artificial intelligence, especially in the domain of LLMs~\cite{wang2022self, heineman2024improving, chen2024alphamath, brown2024large}.
With the development of the VLA paradigm, sampling-and-verification strategies have gradually been introduced in robotics to enhance the performance of VLA models. V-GPS~\cite{nakamoto2024steering} is a pioneering work that selects the highest-ranked actions via offline reinforcement learning to improve robotic manipulation. RoboReward~\cite{lee2026roboreward} constructs a dataset and benchmark for robot reward models. Robomonkey~\cite{kwok2025robomonkey} and Rover~\cite{dai2025rover} train reward models using pretrained VLMs~\cite{llava23} and employ these models to score candidate actions. MG-Select~\cite{jang2025verifier} proposes a non-learning-based approach that directly constructs an action scoring metric using a condition-masking distribution. However, these methods attempt to score actions based solely on image and task inputs, which often results in limited gains. Robotic actions are typically represented as 6D poses, and without effective 3D spatial understanding and reasoning capabilities, it is difficult to distinguish action quality for actions with subtle differences. By incorporating 3D information and enhancing spatial reasoning capabilities, our VeriSpace effectively addresses this limitation.

\section{Method}

\subsection{Test-Time Verification Setup}
We study test-time action selection for vision-language-action (VLA) models in robotic manipulation.
At each time step $t$, a frozen VLA policy $\pi_\theta$ receives the current observation and task instruction, and produces a distribution over executable robot actions.
Instead of committing to a single action prediction in a one-shot manner, we consider a \emph{sample-and-verify} setting, where multiple candidate actions are first proposed and then evaluated before execution.

Given the current observation $O_t = \{I_t^{\text{wrist}}, I_t^{\text{static}}\}$, robot state $S_t $, and task instruction $L$, we sample a set of $k$ candidate actions
\begin{equation}
a_t^{(i)} \sim \pi_\theta(\cdot \mid O_t, S_t, L; \tau), \quad i = 1, \dots, k,
\end{equation}
where $\tau$ is the sampling temperature.
Each candidate action $a_t^{(i)} \in \mathbb{R}^{7}$ is represented as a 7D manipulation command, including the 6-DoF end-effector motion and the gripper action.

The goal of test-time verification is to select, among the sampled candidates, the action most likely to lead to successful task progress.
To this end, we learn an action verifier $\mathcal{F}_\phi(\cdot)$ that assigns a scalar quality score to each candidate action: 
\begin{equation}
s_t^{(i)} = \mathcal{F}_\phi \!\left(a_t^{(i)} \mid I_t^{\text{static}}, D_t, L \right),
\end{equation}
where $D_t$ denotes the depth observation associated with the current scene.
The final action executed by the robot is selected as
\begin{equation}
\widetilde{a}_t = \arg\max_{a_t^{(i)} \in \mathbf{A}_t} \mathcal{F}_\phi \!\left(a_t^{(i)} \mid I_t^{\text{static}}, D_t, L \right),
\end{equation}

The verifier must distinguish among subtly different candidate actions and identify the one that is most spatially valid and most conducive to task completion.

The verifier evaluates candidate actions along two dimensions:
(1) \emph{Action Spatial Reasoning}, which distinguishes fine-grained geometric differences between candidate actions, and
(2) \emph{Goal-Progress Assessment}, which assesses whether an action moves the current state toward successful task completion.

\subsection{VeriSpace: A 3D-Aware Action Verifier}

\begin{figure*}[t]
    \centering{\includegraphics[width=1.0\textwidth]{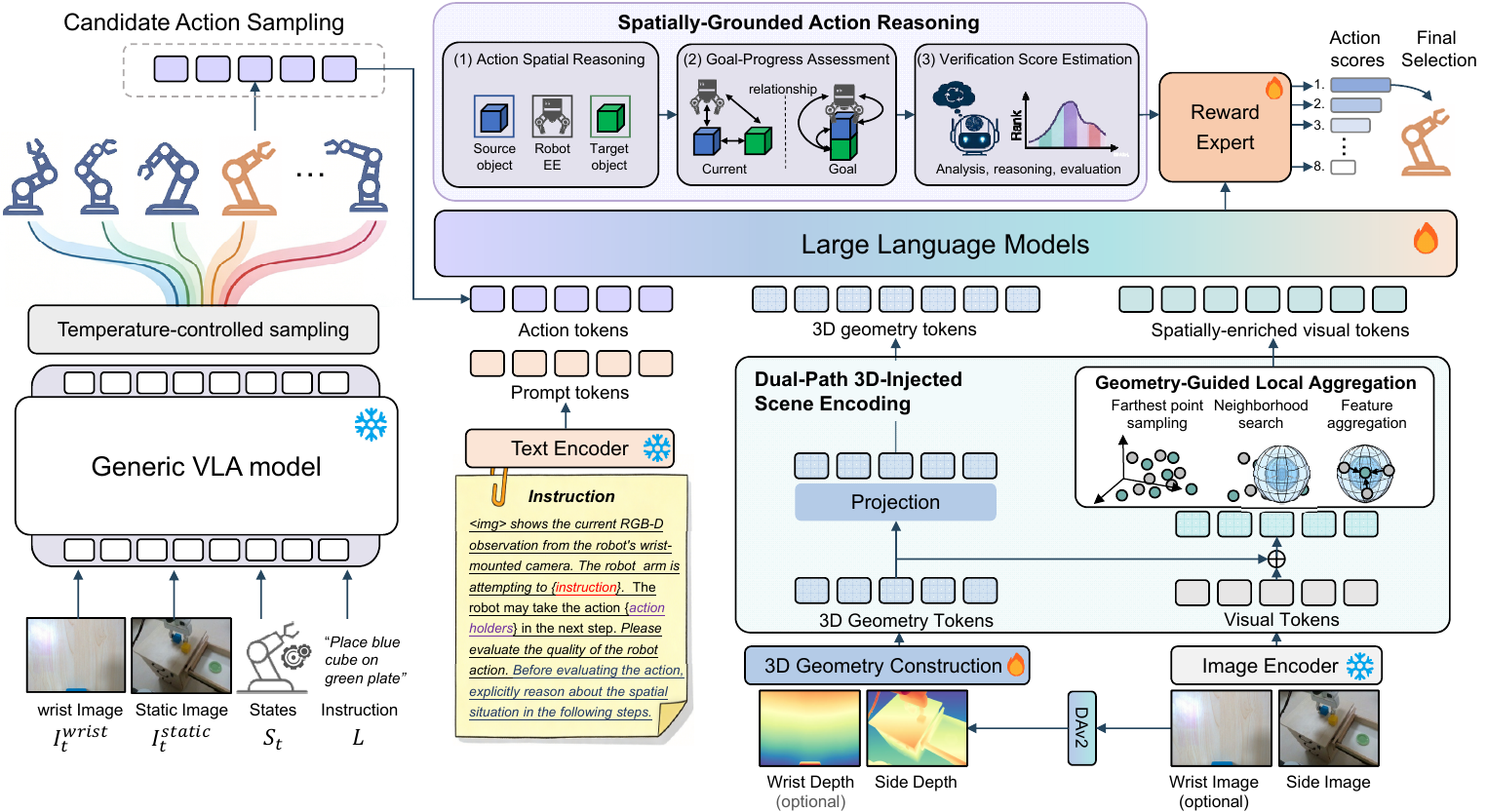}}  
    \vspace{-4mm}
    \caption{
    Overview of VeriSpace.
    At test time, a frozen VLA policy first samples multiple candidate actions from the current observation and instruction.
    VeriSpace then evaluates each candidate through two key components: a 3D-aware scene encoding module that injects explicit scene geometry into the visual representation, and a spatial reasoning module that assesses each action under the current scene and task goal.
    The resulting verification scores are used to rank the sampled actions and select the final action for execution.
    }
    \label{pipeline}
\end{figure*}

As illustrated in Figure~\ref{pipeline}, \textbf{VeriSpace} is a 3D-aware action verifier for evaluating candidate actions in robotic manipulation, built upon a vision-language model (VLM).
Given a candidate action under the current scene and task, VeriSpace performs deliberative reasoning to assess its spatial validity and its contribution to task completion, and outputs a scalar quality score for action selection.
In contrast to prior verifiers that rely primarily on 2D visual features, VeriSpace explicitly incorporates 3D scene structure into the verification process, enabling more reliable discrimination between geometrically similar actions whose outcomes differ due to subtle spatial factors.

Formally, for each sampled candidate action $a_t^{(i)}$, the verifier takes as input the static-view RGB observation $I_t^{\text{static}}$, the associated depth map $D_t$, and the task instruction $L$.
The RGB-D observation is first encoded into a set of spatially informed visual tokens that capture both semantic content and underlying 3D geometry.
These visual tokens are combined with text tokens from the instruction and action tokens representing $a_t^{(i)}$, and jointly processed by the VLM backbone.
Based on the resulting multimodal representations, VeriSpace produces a scalar score $s_t^{(i)}$ that reflects the quality of the candidate action, together with an internal reasoning signal used to guide spatially grounded evaluation.

In this way, VeriSpace functions as a plug-and-play verifier on top of existing VLA policies, re-ranking sampled actions through 3D-aware reasoning without modifying the underlying policy model.

\subsection{Dual-Path 3D-Injected Scene Encoding}
\label{sec:3d_encoding}

To support reliable action verification, the verifier must perceive not only the semantic content of the scene, but also its underlying spatial structure.
To this end, VeriSpace constructs a \textbf{dual-path 3D-injected scene encoding} from the current RGB-D observation, consisting of two complementary token streams: \textbf{3D geometry tokens}, which explicitly represent scene geometry, and \textbf{spatially-enriched visual tokens}, which preserve visual semantics while being grounded in 3D space.
Together, these representations provide geometry-centric and semantics-aware spatial evidence for reasoning. For simplicity, we omit the time index $t$, as all variables refer to the current decision step unless otherwise specified.

\ptitle{3D Geometry Construction.}
Given the static-view RGB observation $I^{\text{static}}$ and the associated depth map $D$, we first recover explicit scene geometry by back-projecting each image location into the 3D workspace using the known camera intrinsic and extrinsic parameters.
This produces a dense 3D coordinate map $P \in \mathbb{R}^{H \times W \times 3}$, which establishes a shared spatial reference for the observed scene.
We then encode the 3D coordinates into the model feature space through a 3D position encoding layer composed of sinusoidal coordinate encoding followed by a learnable two-layer MLP:
\begin{equation}
    P' = \texttt{MLP}(\gamma(P)),
\end{equation}
where $\gamma(\cdot)$ is the sinusoidal encoding function and $P' \in \mathbb{R}^{H \times W \times d}$ denotes the geometry embedding, which forms the basis for both token streams in the verifier.

\ptitle{Dual-Path Scene Tokenization.}
Based on the encoded scene geometry, we construct two complementary token streams for downstream verification.
First, to expose explicit spatial structure to the language model, we project the geometry embedding into the VLM token space to obtain the \textit{3D geometry tokens}
\begin{equation}
    X^{\text{geo}} = \texttt{Projection}(P'),
\end{equation}
where $X^{\text{geo}}$ provides a geometry-centric representation of the scene.
Second, to preserve visual semantics while injecting geometric awareness, we extract visual patch features from the RGB image using a pretrained CLIP encoder~\cite{radford2021learning}, yielding visual tokens $X$.
We then inject the geometry embedding into the visual tokens through element-wise addition:
\begin{equation}
    X^{\text{vis}} = X + P'.
\end{equation}
The resulting \textit{spatially-enriched visual tokens} $X^{\text{vis}}$ retain the strong semantic prior of the original image features while becoming explicitly grounded in 3D space.
In this way, VeriSpace encodes the scene through a dual-path representation: $X^{\text{geo}}$ captures explicit geometric structure, while $X^{\text{vis}}$ captures 
semantics-aware spatial perception.

\ptitle{Geometry-Guided Local Aggregation.}
Although the injected visual tokens $X^{\text{vis}}$ encode the 3D position of each patch, such point-wise grounding alone is insufficient for manipulation-sensitive verification.
For robotic manipulation, action quality often depends on local 3D context---for example, whether a patch lies on an object surface, near an edge, inside a narrow gap, or adjacent to a potential contact or collision region.
To incorporate such local geometric structure into the visual representation, we further perform geometry-guided local aggregation over $X^{\text{vis}}$.

Specifically, we first select a fixed number of seed tokens
\begin{equation}
    \widehat{X}^{\text{vis}} = \texttt{FPS}(X^{\text{vis}}),
\end{equation}
using farthest point sampling (FPS)~\cite{qi2017pointnet++}, where the corresponding 3D coordinates from $P$ are used as the sampling reference.
Let $\widehat{x}^{\text{vis},(j)} \in \widehat{X}^{\text{vis}}$ denote the $j$-th seed token, and let $\widehat{p}^{(j)} \in \mathbb{R}^3$ denote its associated 3D coordinate.
We then retrieve its neighboring visual tokens within a 3D sphere of radius $r$:
\begin{equation}
    \mathcal{N}^{(j)}
    =
    \left\{
    x_{i}^{\text{vis}} \in X^{\text{vis}}
    \;\middle|\;
    \|p_{i} - \widehat{p}^{(j)}\| \le r
    \right\},
\end{equation}
where $p_{i} \in \mathbb{R}^3$ denotes the 3D coordinate of token $x_{i}^{\text{vis}}$.

For each seed token, we aggregate local features from its neighborhood using a geometry-conditioned aggregation function:
\begin{equation}
    \widetilde{x}^{\text{vis},(j)}
    =
    \psi\!\left(\widehat{x}^{\text{vis},(j)}, \mathcal{N}^{(j)}\right)
    =
    \sum_{x_{i}^{\text{vis}} \in \mathcal{N}^{(j)}}
    g\!\left(p_{i} - \widehat{p}^{(j)}\right)\, W x_{i}^{\text{vis}},
\end{equation}
where $g(\cdot)$ is a learnable spatial kernel that maps relative 3D offsets to aggregation weights, and $W$ is a learnable linear projection for token features.
The final aggregated visual token set is denoted by
\begin{equation}
    \widetilde{X}^{\text{vis}}
    =
    \left\{
    \widetilde{x}^{\text{vis},(j)}
    \right\}_{j=1}^{N}.
\end{equation}

This aggregation step injects local 3D neighborhood structure into each visual token, allowing the verifier to capture fine-grained spatial cues that are critical for robotic manipulation.
The resulting dual-path scene tokens, consisting of $\widetilde{X}^{\text{vis}}$ and $X^{\text{geo}}$, are then concatenated with the instruction tokens and candidate action tokens, and jointly fed into the VLM backbone for downstream spatial reasoning and action verification.

\subsection{Spatially-Grounded Action Reasoning}
\label{sec:reasoning}

Given the dual-path scene tokens, VeriSpace evaluates each candidate action through a structured reasoning process that explicitly models both its spatial validity and its contribution to task completion.
The chain-of-thought reasoning proceeds in three steps.

\ptitle{(1) Action Spatial Reasoning.}
VeriSpace first reasons about the fine-grained spatial relationships that determine whether a candidate action is geometrically valid in the current scene.
This includes grounding task-relevant entities, such as the manipulated object, the target object, and the robot end-effector, and analyzing their relative position, orientation, contact, and clearance.
This step enables the verifier to distinguish between actions that appear visually similar but differ in whether they align with the target object, avoid collision, or satisfy local manipulation constraints.

\ptitle{(2) Goal-Progress Assessment.}
VeriSpace then evaluates whether the candidate action moves the current scene toward successful task completion.
This assessment is performed from two complementary perspectives:
\textbf{(2.1) pre-action assessment}, which characterizes the current scene state relative to the task goal, and
\textbf{(2.2) post-action assessment}, which infers the expected scene transition induced by the candidate action and evaluates whether the resulting state is closer to the desired target configuration.
By comparing the pre-action and post-action spatial relations, the verifier estimates how much progress the action makes toward the task objective.

\ptitle{(3) Verification Score Estimation.}
Finally, VeriSpace integrates the above reasoning results to produce the final verification score for the candidate action.
The resulting reasoning-aware representation is mapped to a scalar score, which is used to rank candidate 


\begin{figure}[t]
    \centering{\includegraphics[width=1.0\linewidth]{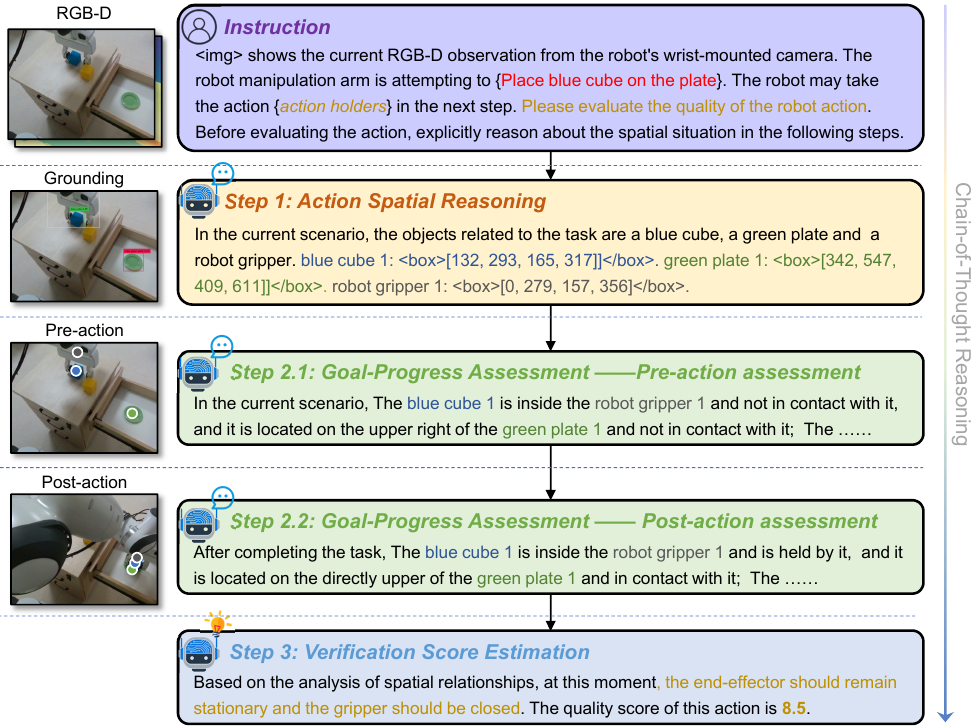}}  
    \vspace{-4mm}
    \caption{
       Spatially-grounded action reasoning process.
       }
    \label{COT}
 \end{figure}

\subsection{Learning Objective}

We train VeriSpace with two objectives: an action evaluation loss $\mathcal{L}_{a}$ for learning candidate action ranking, and a reasoning loss $\mathcal{L}_{\text{CoT}}$ for supervising the reasoning sequence:
\begin{equation}
    \mathcal{L} = \lambda_{a}\mathcal{L}_{a} + \lambda_{\text{CoT}} \mathcal{L}_{\text{CoT}},
\end{equation}
where $\lambda_{a}$ and $\lambda_{\text{CoT}}$ are weighting coefficients.

For action evaluation, we adopt a Bradley--Terry style preference objective over action pairs.
Given a pair of candidate actions $\{a^{W}, a^{L}\}$, where $a^{W}$ leads to better task progress than $a^{L}$, the loss is defined as:
\begin{equation}
   \mathcal{L}_{a} = - \log \sigma\left(
   \mathcal{F}_\phi(a^{W} \mid I^{\text{static}}, D, L)
   -
   \mathcal{F}_\phi(a^{L} \mid I^{\text{static}}, D, L)
   \right),
\end{equation}
where $\sigma(\cdot)$ is the sigmoid function.
This objective encourages the verifier to assign higher scores to actions that are both spatially valid and more conducive to task completion.

For reasoning supervision, we apply a standard cross-entropy loss $\mathcal{L}_{\text{CoT}}$ over the generated CoT sequence, using CoT annotations that capture intermediate spatial reasoning and goal-progress cues.
The details of CoT data construction is provided in Appendix. 

\begin{figure*}[t]
    \centering{\includegraphics[width=1.0\textwidth]{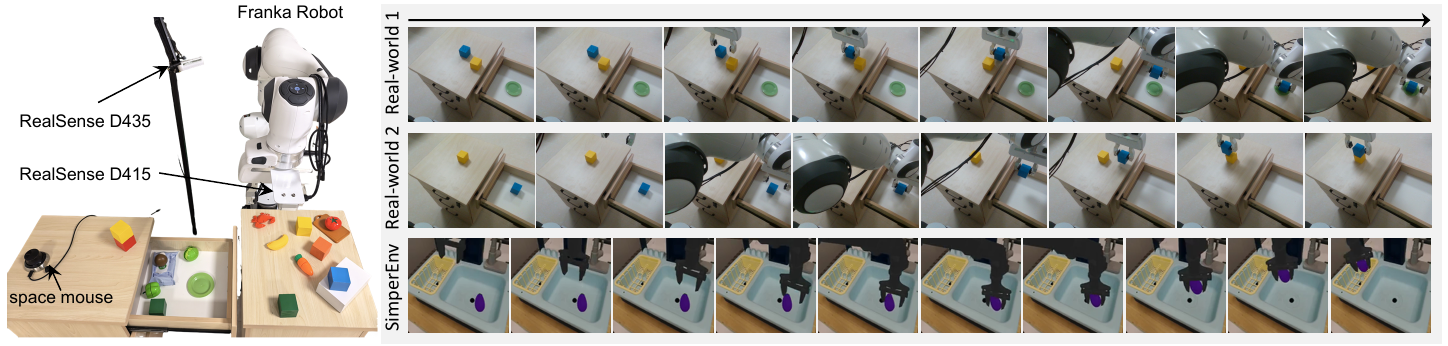}}  
    \vspace{-6mm}
    \caption{
    Real-world setup and qualitative results.
    Left: the robotic platform, including a Franka arm equipped with wrist-mounted and external RGB-D cameras.
    Right: example task executions in real-world  and the SimplerEnv benchmark.
    }
    \label{fig:Qualitative}
 \end{figure*}

\begin{table*}[htbp]
\centering
\setlength{\tabcolsep}{12pt}
\caption{Quantitative comparison on the SimplerEnv-WidowX benchmark~\cite{li24simpler} using OpenVLA~\cite{kim2024openvla} as the baseline. ${\dagger}$ indicates reproduced performance.
We report the average success rate (\%) over 50 trials on 4 tasks.
}
\vspace{-2mm}
\label{tab:SIMPLER}
\resizebox{1.0\linewidth}{!}{
\begin{tabular}{l|llll|l}
\toprule
\multirow{2}{*}{Methods} &
\multicolumn{4}{c|}{Success rate (\%)} &
\multirow{2}{*}{\textbf{Average}} \\
\cmidrule(lr){2-5}
& Eggplant in Basket& Carrot on Plate & Stack Cubes & Spoon on Towel \\
\midrule
OpenVLA~\cite{kim2024openvla} &54.0	&22.0	&28.0	&44.0	&37.0 \\
+V-GPS~\cite{nakamoto2024steering}       &58.0 (+4.0)	&20.0 (-2.0)	&26.0 (-2.0) &40.0  (-4.0) & 36.0(-1.0)\\
+Robomonkey~\cite{kwok2025robomonkey}     &70.0 (+16.0)	&12.0 (-10.0)	&34.0 (+4.0)	& {46.0  (+2.0)} &40.5 (+3.5)\\
+MG-Select~\cite{jang2025verifier}       &62.0 (+8.0)	&16.0 (-6.0)	&34.0 (+4.0)	&44.0 (+0)  & 39.0 (+2.0)\\
\rowcolor{linecolor2}{\textbf{+ VeriSpace (Ours)}}  & \textbf{76.0 (+22.0)} & \textbf{34.0 (+12.0)} & \textbf{62.0 (+34.0)} & \textbf{48.0 (+4.0)} & \textbf{55.0 (+18.0)} \\
\midrule
$\pi_0$-FAST~\cite{black2024pi_0}  &78.0 &66.0 &28.0 &56.0 & 57.0\\
+V-GPS~\cite{nakamoto2024steering}     &78.0 (+0.0)	&60.0 (-6.0)	&18.0 (-10.0) &58.0  (+2.0) & 53.5 (-3.5)\\
+Robomonkey~\cite{kwok2025robomonkey}  &80.0 (+2.0)	&68.0 (+2.0)	&20.0 (-8.0) &\textbf{66.0  (+10.0)} & 58.5 (+1.5)\\
+MG-Select~\cite{jang2025verifier}  &76.0 (-2.0)	&68.0 (+2.0)	&24.0 (-4.0) &{62.0  (+6.0)} & 57.5 (+0.5)\\
\rowcolor{linecolor2}{\textbf{+ VeriSpace (Ours)}} &\textbf{82.0 (+4.0)} & \textbf{72.0 (+6.0)} & \textbf{28.0 (+0.0)} &60.0 (+4.0) &\textbf{60.5 (+3.5)}\\  
\bottomrule 
\end{tabular}
}
\end{table*}

\section{Experiments}

To comprehensively evaluate the performance of VeriSpace, we conduct extensive simulations as well as real-world experiments on physical robots. We design a series of experiments to address the following questions: (1) How does VeriSpace perform on the SimplerEnv-WidowX~\cite{li24simpler} and LIBERO benchmarks
compared with state-of-the-art test-time methods~\cite{kwok2025robomonkey, jang2025verifier}?  (2) How does VeriSpace perform on real robots? (3) What is the generalization ability and applicability of our method across different VLA models? (4)~What are the contributions of each component of our approach?

\subsection{Experimental Setups}

\ptitle{Implementation Details.}
We train our model using 8 NVIDIA A800 GPUs for a total of 12 days. The batch size is set to 16 per GPU, resulting in an overall batch size of 128, and the learning rate is $2 \times 10^{-5}$. For the verifier, we adopt LLaVA-7B~\cite{llava23} as the underlying VLM and fine-tune it using LoRA~\cite{hu2022lora}, with $r=512$ and $ \alpha = 128$. Our model takes as input third-person RGB-D frames with a resolution of $224 \times 224$, along with a task instruction and a candidate action, and outputs a score for the candidate action. Additional implementation details are provided in the Appendix.


\ptitle{Evaluation Baselines.}
We adopt OpenVLA~\cite{kim2024openvla} and $\pi_0$-FAST~\cite{black2024pi_0} as VLA baselines and reproduce their performance across different benchmarks. We compare our method with state-of-the-art test-time verification strategies, including V-GPS~\cite{nakamoto2024steering}, Robomonkey~\cite{kwok2025robomonkey}, and MG-Select~\cite{jang2025verifier}. To ensure a fair comparison, we standardize the number of sampled candidate actions to 8 and set the sampling temperature to 1, while keeping all other parameters consistent with the official implementations.

\begin{table*}[htbp]
\caption{
Quantitative comparison on the LIBERO benchmark.
Success rates (\%) are averaged over 50 trials per task across four task suites, each containing 10 tasks.
}
\vspace{-2mm}
\setlength{\tabcolsep}{12pt}
\begin{center}
\resizebox{1.0\textwidth}{!}{
\begin{tabular}{l|llll|l}
\toprule
 \multirow{1}{*}{Model} & LIBERO-Spatial & LIBERO-Object & LIBERO-Goal & LIBERO-Long & \multirow{1}{*}{\textbf{Average}} \\
\midrule
OpenVLA~\cite{kim2024openvla} & {84.7} & 65.2  & {74.4} & 53.1 & 69.3 \\
{+ MG-Select}~\cite{jang2025verifier}    & 81.7 (-3.0) & 72.5 (+7.3) & 73.6 (-0.8) & \textbf{55.4 (+2.3)} & 70.8 (+1.5) \\
\rowcolor{linecolor2}{\textbf{+ VeriSpace (Ours)}}   & \textbf{93.0 (+8.3)} & \textbf{75.9 (+10.7)} & \textbf{78.1 (+3.7)} & 53.7 (+0.6) & \textbf{75.2 (+5.9)} \\
\midrule
$\pi_0$-FAST~\cite{black2024pi_0} & {92.2} & 93.7  & {91.8} & 75.0 & 88.1\\
{+ MG-Select}~\cite{jang2025verifier}   & 97.2 (+5.0) & 98.0 (+4.3) & 94.5 (+2.7) & \textbf{82.7 (+7.7)} & 93.1 (+5.0) \\
\rowcolor{linecolor2}{\textbf{+ VeriSpace (Ours)}}  & \textbf{98.4 (+6.2)} & \textbf{98.2 (+4.5)} & \textbf{98.1 (+6.3)} & 82.1 (+7.1) & \textbf{94.2 (+6.1)} \\
\bottomrule
\end{tabular}}
\label{tab:libero_main}
\end{center}
\end{table*}

\begin{table*}[htbp]
\caption{Quantitative results on the real-world experiments. We report average success rates (\%) over 50 trials for each task, including both in-distribution and out-of-distribution settings.}
\vspace{-2mm}
\begin{center}
\resizebox{1.0\textwidth}{!}{
\begin{tabular}{llllllll}
\toprule
 \multirow{2}{*}{Model} & 
 \multicolumn{4}{c}{In-distribution} &  \multicolumn{2}{c}{Out-of-distribution}
 & \multirow{2}{*}{\textbf{Average}} \\
 \cmidrule(lr){2-5}
  \cmidrule(lr){6-7}
 & Tomato on Plate & Cube on Plate & Stack Cubes & Pepper on Plate &  Cube in Drawer & Banana on Plate  & \\
\midrule
$\pi_0$-FAST~\cite{black2024pi_0}  &60.0 &64.0 &54.0 &60.0 &54.0 &52.0 &58.3\\
\rowcolor{linecolor2}{\textbf{+ VeriSpace}}  & \textbf{72.0 (+12.0)} & \textbf{82.0 (+18.0)} & \textbf{74.0 (+20.0)} & \textbf{64.0 (+4.0)} & \textbf{76.0 (+22.0)}  &\textbf{66.0 (+14.0)} &\textbf{72.3 (+14.0)}\\
\bottomrule
\end{tabular}}
\label{tab:Real}
\end{center}
\end{table*}

\subsection{Simulation Evaluation on SimplerEnv}

The SimplerEnv-WidowX benchmark~\cite{li24simpler} consists of 4 pick-and-place tasks, which are used to evaluate whether our method improves task completion robustness in simulation. As it does not provide a training set, following~\cite{qu2025spatialvla, kwok2025robomonkey}, we train our model on the BridgeData V2 dataset~\cite{walke2023bridgedata, kwok2025robomonkey}. 

\ptitle{Evaluation Results with OpenVLA.}
The upper portion of Table~\ref{tab:SIMPLER} presents the performance of VeriSpace when integrated with OpenVLA~\cite{kim2024openvla}. On the SimplerEnv-WidowX benchmark~\cite{li24simpler}, our method achieves substantial improvements over OpenVLA across all tasks, with an average gain of +18.0 pp, reaching 55.0\%. Notably, on the ``Stack Cubes'' task, which demands stronger spatial reasoning, our method outperforms the vanilla VLA baseline~\cite{kim2024openvla} by 34 pp. Compared to the state-of-the-art Robomonkey~\cite{kwok2025robomonkey}, VeriSpace achieves an additional improvement of 14.5 pp. These results demonstrate that our approach significantly enhances the robustness of robotic manipulation and improves success rates in spatially fine-grained tasks through explicit spatial reasoning.

\ptitle{Evaluation Results with $\pi_0$-FAST.}
The lower portion of Table~\ref{tab:SIMPLER} reports the results when integrating VeriSpace with $\pi_0$-FAST~\cite{black2024pi_0}. Our method consistently improves performance over the baseline, achieving an average success rate of 60.5\% (+3.5 pp). Gains are observed across most tasks, with notable improvements on \textit{Carrot on Plate} (+6 pp) and \textit{Eggplant in Basket} (+4 pp), while maintaining performance on Stack Cubes. Compared to prior methods such as Robomonkey~\cite{kwok2025robomonkey}, VeriSpace achieves higher overall performance (+2.0 pp), demonstrating its effectiveness even when combined with a stronger policy backbone. These results suggest that VeriSpace provides complementary benefits by enhancing spatial verification, leading to more reliable manipulation outcomes.

\subsection{Simulation Evaluation on LIBERO}

The LIBERO benchmark~\cite{liu2023libero} comprises a diverse set of long-horizon manipulation tasks designed to evaluate the generalization and compositional reasoning capabilities. It is divided into four subsets: Spatial, Object, Goal, Long. For training, we finetune our model on the LIBERO training dataset. 

\ptitle{Evaluation Results.}
Table~\ref{tab:libero_main} summarizes the performance of VeriSpace with both OpenVLA~\cite{kim2024openvla} and $\pi_0$-FAST~\cite{black2024pi_0}. When combined with OpenVLA, our method achieves an overall success rate of 75.2\%, yielding improvements of +5.9 pp over the baseline and +4.4 pp over MG-Select~\cite{jang2025verifier}. Notably, VeriSpace attains 93.0\% on LIBERO-Spatial, highlighting its strength in spatial manipulation, while gains on long-horizon tasks remain relatively modest due to increased reasoning complexity.
When applied to $\pi_0$-FAST, VeriSpace continues to deliver consistent improvements, boosting overall performance by +6.1 pp and surpassing the current SOTA~\cite{jang2025verifier} by +1.1 pp. These results demonstrate the strong generalizability of our approach and its effectiveness across different VLA backbones.



\subsection{Real-World Evaluation}

\ptitle{Real-World Benchmarks and Settings.}
To evaluate the effectiveness of VeriSpace in real-world scenarios, we conduct experiments using a Franka robot. The experimental setup is illustrated in Figure~\ref{fig:Qualitative}, with a more detailed description of the real-world settings provided in the appendix.
For benchmarks, we design both in-distribution and out-of-distribution tasks. The in-distribution tasks consist of four pick-and-place tasks with spatial variations. For the training set, each task includes 50 demonstrations, with an average trajectory length of approximately 500 steps. The out-of-distribution tasks involve previously unseen objects and spatial configurations (different heights and positions), comprising two tasks in total, and are used to evaluate the zero-shot generalization capability of our method. 

\ptitle{In-distribution Tasks.}
We first evaluate our VeriSpace on in-distribution tasks, as shown in Table~\ref{tab:Real}. When combined with $\pi_0$-FAST, our method achieves improvements across all tasks, yielding an average success rate of 73.0\%, corresponding to a +13.0 pp gain over the baseline.
In particular, VeriSpace improves performance by +18.0 pp on \textit{Cube on Plate} and +20.0 pp on \textit{Stack Cubes}, indicating its strong capability in handling precise manipulation tasks. In addition, as shown in Figure~\ref{fig:Qualitative} (Real-World 1 and 2), our method remains robust in scenarios involving complex spatial relationships, further demonstrating its superiority in such settings. These results demonstrate that VeriSpace effectively enhances the robustness of robot manipulation under in-distribution tasks.

\ptitle{Out-of-distribution Tasks.}
We further evaluate the generalization ability of VeriSpace on out-of-distribution (OOD) tasks. As shown in Table~\ref{tab:Real}, our method achieves significant improvements, increasing the average success rate from 53.0\% to 71.0\%.
Notably, VeriSpace improves performance by +22.0 pp on \textit{Cube in Drawer} and +14.0 pp on \textit{Banana on Plate}, both of which involve novel object configurations and unseen spatial arrangements. The larger performance gains compared to in-distribution tasks suggest that our method is particularly effective in mitigating distribution shifts, likely due to its ability to verify action candidates based on spatial understanding. These results highlight the robustness and generalization capability of VeriSpace in real-world scenarios.

\begin{table*}[htbp]
\caption{
Results on the SimplerEnv-WidowX benchmark across different VLA backbones. 
Task success rates (\%) are shown for each base policy and its corresponding VeriSpace-enhanced version.
}
\vspace{-2mm}
\setlength{\tabcolsep}{12pt}
\begin{center}
\resizebox{1.0\textwidth}{!}{
\begin{tabular}{l|llll|l}
\toprule
 \multirow{1}{*}{Model} & Eggplant in Basket & Carrot on Plate & Stack Cubes & Spoon on Towel & \multirow{1}{*}{\textbf{Average}} \\
 \midrule
Octo~\cite{team2024octo}  & 56.9 & 9.7 & 4.2 & 47.2 & 29.5 \\
\rowcolor{linecolor2}{\textbf{+ VeriSpace (Ours)}}  & \textbf{68.0 (+11.1)} & \textbf{26.0 (+16.3)} & \textbf{12.0 (+7.8)} & \textbf{52.0 (+4.8)}  & \textbf{39.5 (+10.0)} \\
\midrule
OpenVLA~\cite{kim2024openvla} &54.0	&22.0	&28.0	&44.0	&37.0 \\
\rowcolor{linecolor2}{\textbf{+ VeriSpace (Ours)}}  & \textbf{76.0 (+22.0)} & \textbf{34.0 (+12.0)} & \textbf{62.0 (+34.0)} & \textbf{48.0 (+4.0)} & \textbf{55.0 (+18.0)} \\
\midrule
$\pi_0$-FAST~\cite{black2024pi_0}  &78.0 &66.0 &28.0 &56.0 & 57.0\\
\rowcolor{linecolor2}{\textbf{+ VeriSpace (Ours)}} &\textbf{82.0 (+4.0)} & \textbf{72.0 (+6.0)} & {28.0 (+0.0)} &\textbf{60.0 (+4.0)} &\textbf{60.5 (+3.5)}\\ 
\midrule
SpatialVLA~\cite{qu2025spatialvla} &100.0 &25.0 &29.2 &16.7 &42.7\\
\rowcolor{linecolor2}{\textbf{+ VeriSpace (Ours)}}  &{100.0 (+0.0)} & \textbf{36.0 (+11.0)} & \textbf{48.0 (+18.8)} &\textbf{28.0 (+11.3)} &\textbf{53.0 (+10.3)}\\ 
\bottomrule
\end{tabular}}
\label{tab:Enhancement}
\end{center}
\end{table*}

\subsection{Applicability across Different VLAs}
To validate the broad applicability and strong generalization capability of VeriSpace, we conduct compatibility experiments on the SimplerEnv-WidowX benchmark~\cite{li24simpler}. Specifically, we integrate our method with several VLA models, including OpenVLA~\cite{kim2024openvla}, $\pi_0$-FAST~\cite{black2024pi_0}, SpatialVLA~\cite{qu2025spatialvla}, and Octo~\cite{team2024octo}. The results, summarized in Table~\ref{tab:Enhancement}, show that combining VeriSpace with different VLA models consistently improves the average success rate by $3 \sim 18$ pp.
In particular, OpenVLA~\cite{kim2024openvla} benefits significantly from our method, achieving an improvement of 18 pp over its baseline performance. These results demonstrate the strong adaptability and generalization ability of VeriSpace across diverse models. Notably, $\pi_0$ + VeriSpace achieves a highest success rate of 60.5\%.


\begin{table}[htbp]
    \caption{Ablation study on SimplerEnv and real-world tasks. }
    \vspace{-2mm}
    \centering
    \scriptsize
    \resizebox{1.0\linewidth}{!}{
    \begin{tabular}{l|l|cc}
    \toprule
    No.  &Methods        & SimplerEnv  & Real-world \\
    \midrule
    1)    &w/o 3D input  &42.0 & 63.6 \\
    \rowcolor{linecolor2}{2)}         &w 3D input   &\textbf{55.0}  &\textbf{72.3} \\
    \midrule
    3)         &w/o D3SE   &46.5 & 68.0 \\
    \rowcolor{linecolor2}{4)}         &w D3SE  &\textbf{55.0}  &\textbf{72.3} \\
    \midrule
    5)         &FPS Agg.   &52.0 &70.3   \\
    6)         &Voxelization Agg.   &50.5 &70.6   \\
    \rowcolor{linecolor2}{7)}         &GLA &\textbf{55.0}  &\textbf{72.3} \\
    \midrule
    8)         &w/o SGAR   &48.5 &67.3   \\
    \rowcolor{linecolor2}{9)}         &w SGAR  &\textbf{55.0}  &\textbf{72.3} \\
    \bottomrule
    \end{tabular}
    }
    \label{Ablation}
\end{table}

\subsection{Ablation Studies}

\ptitle{3D Information Input.}
3D input is crucial for spatial reasoning in action scoring. We conduct an ablation study by removing depth input. As shown in Table~\ref{Ablation} (1) and (2), without 3D input, our method achieves only 42.0\% on the SimplerEnv benchmark, representing a drop of 13.0 pp. This result highlights the importance of 3D information for accurate action evaluation.

\ptitle{Dual-Path 3D-Injected Scene Encoding.}
As shown in Table~\ref{Ablation} (3) and (4), we remove the dual-path 3D-injected scene encoding (D3SE). Without D3SE, no 3D positional encoding is incorporated, and the model operates on standard 2D tokens rather than spatial-aware 2D tokens. The results indicate removing this module leads to consistent performance degradation across both benchmarks.

\ptitle{Geometry-Guided Local Aggregation.}
We replace our geometry-guided local aggregation (GLA) with FPS aggregation and voxelization aggregation. The results, shown in Table~\ref{Ablation} (5), (6), and (7), demonstrate that FPS and voxelization aggregation achieve similar performance, but both underperform our spherical neighborhood pooling by approximately 4 pp. This is because GLA enhances the model’s ability to capture local geometric structures during token sampling, thereby reducing information loss.

\ptitle{Spatially-Grounded Action Reasoning.}
We further conduct an ablation by removing the spatially-grounded action reasoning (SGAR) process and directly predicting action scores. As shown in Table~\ref{Ablation} (8) and (9), the success rate on SimplerEnv drops significantly from 55.0\% to 48.5\%. This demonstrates that explicit spatial reasoning plays a critical role in accurate action scoring.

\begin{figure}[t]
    \centering{\includegraphics[width=1.0\linewidth]{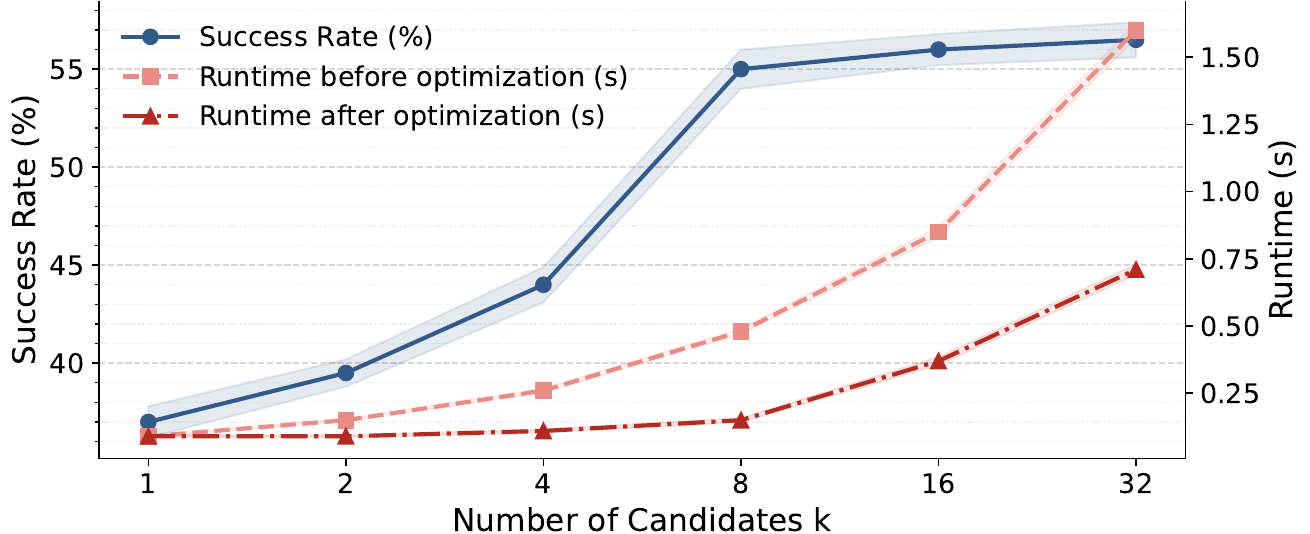}}  
    \vspace{-2mm}
    \caption{
       Success rate and inference runtime as the number of action candidates $k$ increases.
       }
    \label{linechart}
\end{figure}

\ptitle{Number of Action Candidates $k$.}
Figure~\ref{linechart} shows the task success rate and per-step inference runtime on the SimplerEnv benchmark under different candidate set sizes $k$.
As $k$ increases, the success rate improves consistently, rising from 37.0\% at $k{=}1$ to 55.0\% at $k{=}8$, and reaching 56.0\% at $k{=}32$, indicating that test-time candidate scaling provides clear gains for action selection.
At the same time, larger candidate sets inevitably introduce additional verification cost.
In practice, however, this overhead remains well controlled through parallel action verification across multiple GPUs and KV-cache reuse~\cite{kwok2025robomonkey}.
As shown in Figure~\ref{linechart}, the optimized runtime is substantially lower than the unoptimized baseline across all $k$, making larger candidate sets practical at test time. See Appendix~\ref{Runtime} for a more detailed runtime analysis.

\section{Conclusion}
We presented VeriSpace, a 3D-aware action verifier for improving test-time decision-making in vision-language-action systems. By replacing direct action scoring with geometry-aware perception and spatially grounded reasoning, VeriSpace enables more reliable discrimination between subtle but outcome-critical manipulation decisions. Our results show that robust robotic behavior depends not only on proposing plausible actions, but on evaluating them within a spatially informed and task-aware generate-and-verify loop before execution. Taken together, these findings point to a practical path toward more reliable and generalizable robotic foundation models, where action generation is coupled with explicit spatial understanding and decision-time reasoning.



\appendix

\section{Implementation Details}

\subsection{VeriSpace Architecture}

\ptitle{Text Encoder.}
We leverage the tokenizer and text embedding layers inherited from LLaVA-7B~\cite{llava23} to process natural language inputs. Specifically, textual instructions are tokenized into subword units and mapped into dense embeddings that capture semantic and syntactic information. These embeddings are directly fed into the language model backbone, enabling seamless integration with visual tokens and facilitating joint multimodal reasoning.

\ptitle{Image Encoder.}
We adopt a CLIP-pretrained Vision Transformer (ViT-L/14)~\cite{radford2021learning} as the image encoder in VeriSpace. Given an input image, it is first resized (e.g., to $336 \times 336$) and partitioned into non-overlapping patches, each of which is linearly projected into a latent representation. This process produces a sequence of visual tokens (e.g., $24 \times 24 = 576$), optionally including a global representation token. To bridge the modality gap, these visual features are further projected into the language embedding space via a learned linear projector, enabling effective alignment with textual representations.

\ptitle{Action Tokenizer.}
We use an action tokenizer~\cite{kim2024openvla, kwok2025robomonkey} to discretize continuous robot actions into a finite set of tokens compatible with the language model vocabulary. 
Specifically, each action dimension is uniformly quantized into a fixed number of bins within a predefined range $[a_{\min}, a_{\max}]$. 
The resulting discretized values are then mapped to a subset of rarely used tokens located at the tail of the tokenizer vocabulary. 
This design allows action representations to be seamlessly integrated into the language modeling framework while avoiding interference with frequently used textual tokens.

\ptitle{Large Language Model.}
We employ LLaVA-7B (Vicuna-7B)~\cite{llava23} as the language backbone of VeriSpace. Vicuna-7B is a decoder-only Transformer model with billions of parameters, derived from LLaMA and further fine-tuned on conversational data to enhance instruction-following and dialogue capabilities. It operates in an autoregressive manner, modeling the probability of each token conditioned on previous tokens and multimodal inputs. By jointly attending to both visual and textual embeddings, the model is capable of performing complex multimodal reasoning and generating coherent responses grounded in visual context.




\ptitle{Reward Expert.}
We implement the reward expert as a lightweight MLP. 
It takes as input the final hidden state produced by the large language model and maps it to a scalar score $s_t^{(i)}$. This score $s_t^{(i)}$ represents the final quality assessment of the candidate action $a_t^{(i)}$.

\begin{figure}[htbp]
    \centering{\includegraphics[width=1.0\linewidth]{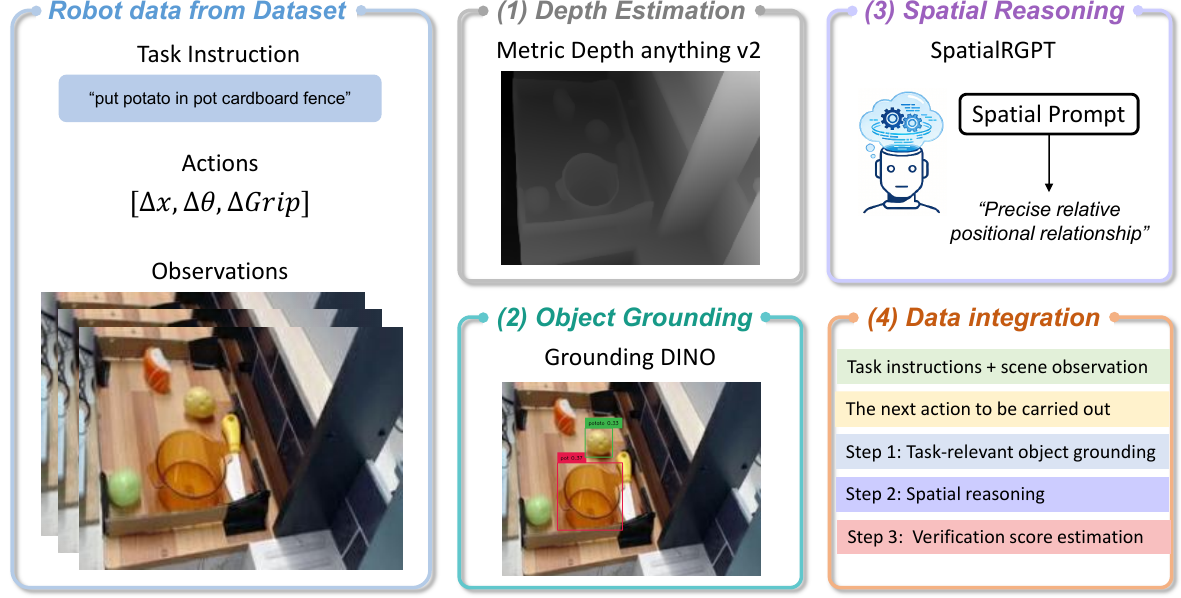}}  
    \caption{
    Our pipeline for generating synthetic embodied chain-of-thought data.
    }
    \label{qr:pipeline}
\end{figure}

\subsection{Dataset Construction}
We illustrate our pipeline for generating synthetic embodied chain-of-thought data in Figure~\ref{qr:pipeline}. 
Given a robotic dataset, we take as input the task instruction, robot actions, and current observations. First, we estimate depth information from the current observation using Metric Depth Anything V2~\cite{yang2024depthv2}, which serves as the 3D input. Next, we leverage Grounding DINO~\cite{liu2024grounding} to extract relevant objects with the task instruction and obtain their corresponding bounding boxes. The estimated depth and localization information, together with the original observation and task instruction, are then fed into SpatialRGPT~\cite{cheng2024spatialrgpt} to reason about the relative spatial relationships between objects in the current and goal states. Finally, all intermediate outputs are integrated to construct a synthetic chain-of-thought for embodied reasoning.

\ptitle{Preference Label.}
Following~\cite{kwok2025robomonkey}, we generate $\binom{N}{2}$ preference pairs per timestep by sampling $N$ candidates from the Generic VLA model. We then compute the RMSE against the ground-truth action. The action with the lower error receives the preferred label (1), and the other (0).

\begin{figure*}[htbp]
    \centering{\includegraphics[width=1.0\textwidth]{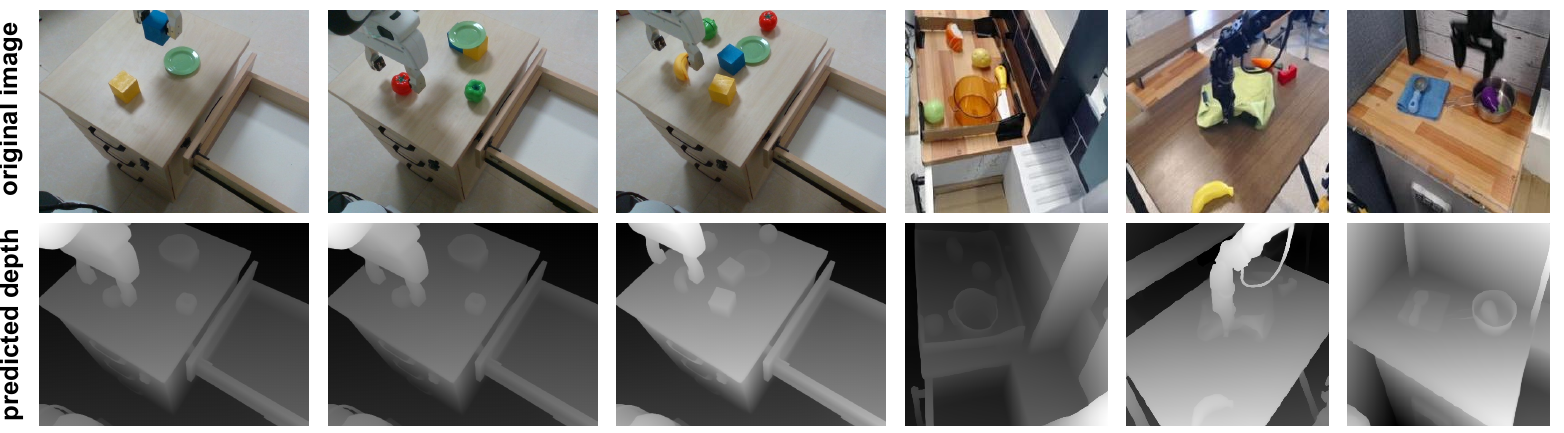}}  
    \caption{
    Visualization results of depth estimation.
    }
    \label{qr:depth}
\end{figure*}

\ptitle{Depth Estimation.}
As described in the main paper, we adopt Metric Depth Anything V2~\cite{yang2024depthv2} as our metric depth estimator to recover 3D geometric information from monocular observations. We find that this model produces stable and accurate depth predictions across diverse real-world and simulation environments, which is crucial for downstream spatial reasoning. Examples of depth estimation on the robot dataset is shown in Figure~\ref{qr:depth}.
Given an input RGB observation, the depth estimator predicts a dense depth map that serves as a proxy for the scene geometry. This depth information is further utilized to infer relative spatial relationships between objects, complementing the 2D visual features. In particular, the recovered geometry provides important cues for estimating object positions, distances, and height differences, which are essential for manipulation tasks involving spatial constraints.

\begin{figure*}[htbp]
    \centering{\includegraphics[width=1.0\textwidth]{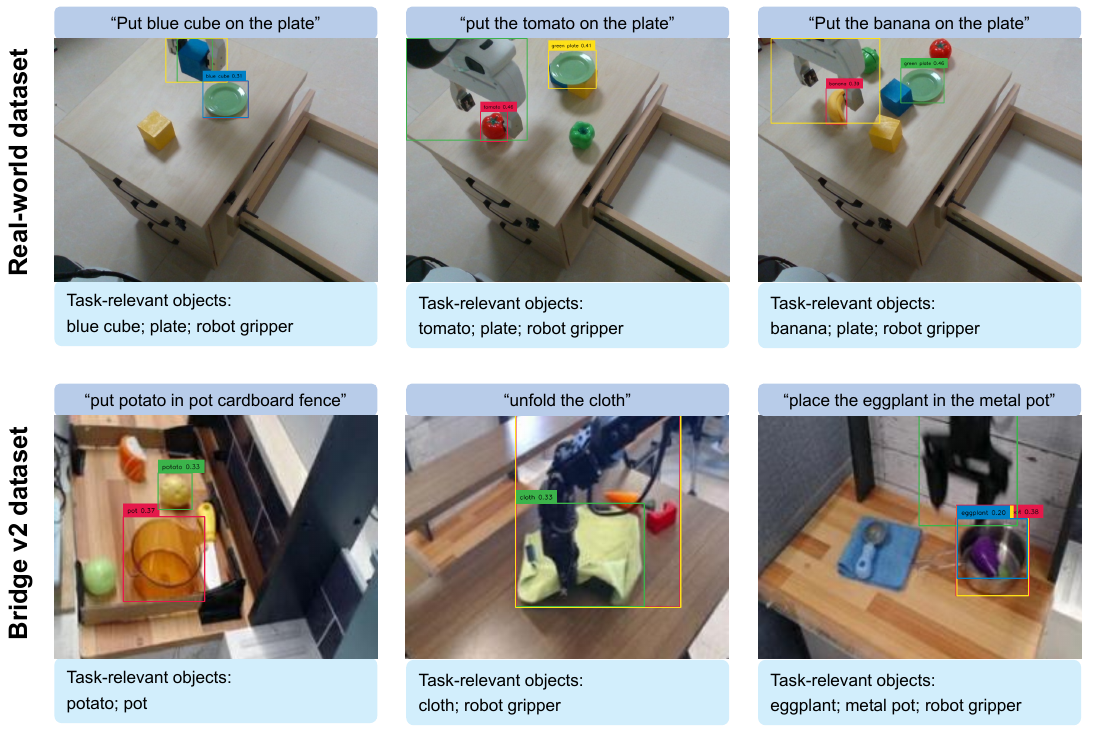}}  
    \caption{
    Visualization results of task-relevant objects grounding.
    }
    \label{qr:grounding}
\end{figure*}

\ptitle{Task-Relevant Objects Grounding.}
To identify objects relevant to the task, we leverage Grounding DINO~\cite{liu2024grounding} to perform language-conditioned object detection. Given a task instruction, we first extract key object phrases and use them as textual queries to localize corresponding regions in the image. The model outputs bounding boxes associated with the queried objects. The results of spatial localization is shown in Figure~\ref{qr:grounding}.
These detected bounding boxes are further aligned with the estimated depth map to obtain approximate 3D positions of task-relevant objects. 

\begin{figure*}[htbp]
    \centering{\includegraphics[width=1.0\textwidth]{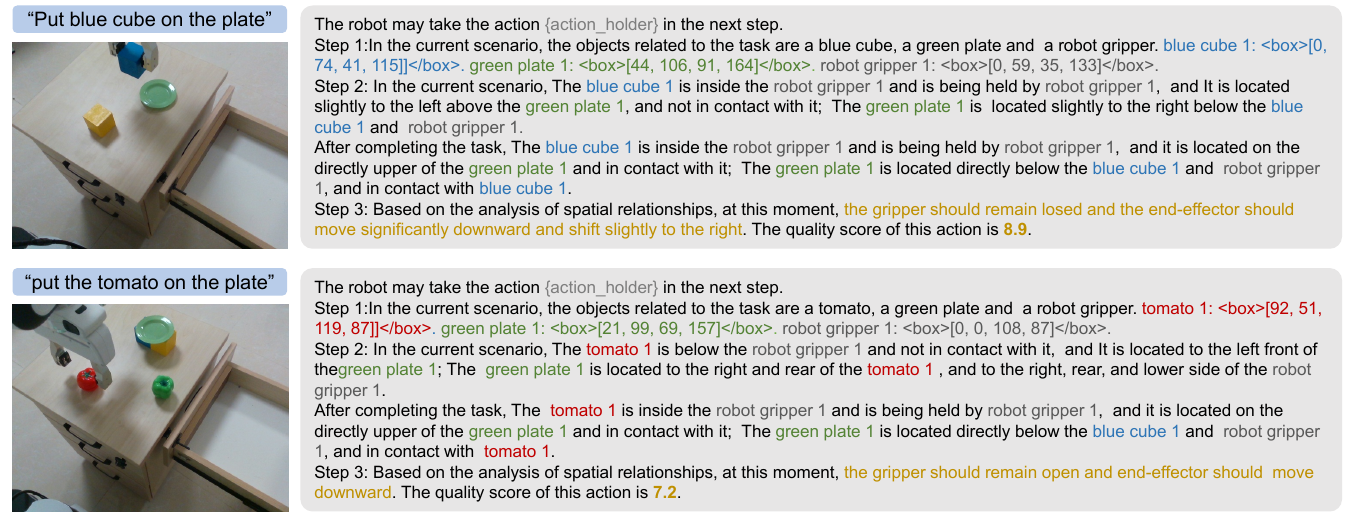}}  
    \caption{
    Examples of CoT data annotation.
    }
    \label{qr:cot}
\end{figure*}

\ptitle{Spatial Relationship Reasoning.}
Building upon the grounded object representations and depth cues, we employ SpatialRGPT-8B~\cite{cheng2024spatialrgpt} to perform spatial reasoning over the scene. The model takes as input the original observation, task instruction, object bounding boxes, and corresponding depth information, and infers the relative spatial relationships among objects in both the current and target states.
Specifically, this model will reason about the fine-grained relative positional relationships of task-related objects. This reasoning process produces structured descriptions among task-related objects, which serve as intermediate supervision for constructing chain-of-thought data. The outputs of spatial relationship reasoning are illustrated in Figure~\ref{qr:cot}.

\ptitle{Chain-of-Thought Data Example.}
Finally, we integrate all intermediate outputs, including depth estimation, object grounding, and spatial reasoning results, to construct synthetic embodied chain-of-thought data. Each CoT instance consists of step-by-step reasoning that connects the current observation to the desired goal state, along with the corresponding action.
Figure~\ref{qr:cot} presents examples of the final annotated CoT data.

The prompt used for action verification is illustrated in Figure~\ref{fig:prompt}. It consists of the task instruction and visual observations, followed by a textual prompt that guides the model to perform chain-of-thought (CoT) reasoning. The model outputs both the intermediate reasoning process and the final action score prediction.

\begin{figure*}[t]
\centering
\small
\begin{tcolorbox}[colback=gray!5, colframe=gray!50]
USER: <img> shows the current observation from the robot's third-person camera.

The robot manipulation arm is attempting to <instruction>. What action should the robot take to effectively accomplish the task?

ASSISTANT: The robot may take the action <action\_holder> in the next step.

USER: Please evaluate the quality of the robot action.

Before evaluating the action, explicitly reason about the spatial situation and the contribution to task completion in the following steps:

Step 1: Task-relevant object grounding. 

Identify whether the target object for the task is visible in the scene and describe its approximate location.

Step 2: Spatial reasoning and goal-progress assessment.

Using the image and depth information, describe the relative positional relationship between the task-related objects in the current state and the goal state.

Step 3: Verification score estimation.

Based on the above reasoning, evaluate the quality of the robot action. You should also consider the following factors: task success, spatial relationship between the objects and the robot, collision risk, human preferences, and contribution to task completion.

ASSISTANT:

USER: Please output the final quality score the robot action.

ASSISTANT: The quality score of the robot action is
\end{tcolorbox}
\caption{Prompt template for candidate action scoring.}
\label{fig:prompt}
\end{figure*}



\subsection{Additional Training Details}

\ptitle{Detailed Loss Function.}
We train VeriSpace with two objectives: an action evaluation loss $\mathcal{L}_{a}$ for learning candidate action ranking, and a reasoning loss $\mathcal{L}_{\text{CoT}}$ for supervising the reasoning sequence:
\begin{equation}
    \mathcal{L} = \lambda_{a}\mathcal{L}_{a} + \lambda_{\text{CoT}} \mathcal{L}_{\text{CoT}},
\end{equation}
where $\lambda_{a}$ and $\lambda_{\text{CoT}}$ are weighting coefficients.

For action evaluation, we adopt a Bradley--Terry style preference objective over action pairs.
Given a pair of candidate actions $\{a^{W}, a^{L}\}$, where $a^{W}$ leads to better task progress than $a^{L}$, the loss is defined as:
\begin{equation}
   \mathcal{L}_{a} = - \log \sigma\left(
   \mathcal{F}_\phi(a^{W} \mid I^{\text{static}}, D, L)
   -
   \mathcal{F}_\phi(a^{L} \mid I^{\text{static}}, D, L)
   \right),
\end{equation}
where $\sigma(\cdot)$ is the sigmoid function.
This objective encourages the verifier to assign higher scores to actions that are both spatially valid and more conducive to task completion.

For reasoning supervision, we apply a standard cross-entropy loss $\mathcal{L}_{\text{CoT}}$ over the generated CoT sequence, using CoT annotations that capture intermediate spatial reasoning and goal-progress cues. The final CoT cross-entropy loss $\mathcal{L}_{CoT}$ is as follows:
\begin{equation}
\mathcal{L}_{CoT} = - \sum_{t=1}^{T} \log P_\phi(y_t \mid y_{<t}, L')
\end{equation}
where $L'$ denotes the instruction prompt, 
$y = (y_1, y_2, \dots, y_T)$ denotes the complete CoT sequence including both intermediate reasoning steps, 
$y_{<t}$ denotes all tokens preceding step $t$, 
$P_\phi$ denotes the probability distribution parameterized by the model, 
and $T$ denotes the total length of the CoT sequence.

\ptitle{Hyperparameter Setting.}
We train our model using 8 NVIDIA A800 GPUs for a total of 12 days. The batch size is set to 16 per GPU, resulting in an overall batch size of 128, and the learning rate is $2 \times 10^{-5}$. For the verifier, we adopt LLaVA-7B~\cite{llava23} as the underlying VLM and fine-tune it using LoRA~\cite{hu2022lora}, with $r=512$ and $ \alpha = 128$. Our model takes as input third-person RGB-D frames with a resolution of $224 \times 224$, along with a task instruction and a candidate action, and outputs a score for the candidate action. Detailed training configurations are provided in Table~\ref{tab:train_config}.
\begin{table}[h]
\centering
\caption{Hyperparameter setting.}
\label{tab:train_config}
\begin{tabular}{l l}
\toprule
\textbf{Hyperparameters} & \textbf{Value} \\
\midrule
\# GPUs & 8 \\
Batch size & 16 / GPU (128 effective) \\
Learning rate (LR) & $2 \times 10^{-5}$ \\
LR Schedule & Constant \\
Weight decay & 0.0 \\
Optimizer & AdamW \\
Epochs & 1 \\
Warm-up ratio & 0.003\\
Warm-up schedule & Constant with warmup \\
LoRA $r$ & 512 \\
LoRA $\alpha$  & 128 \\
bits & 16 \\
Number of candidate actions $k$ & 8 \\
\bottomrule
\end{tabular}
\end{table}

\section{Experiments}

\subsection{Additional Simulation Details}

\ptitle{ SimplerEnv-WidowX .}
The SimplerEnv-WidowX benchmark~\cite{li24simpler} consists of four representative pick-and-place tasks designed to evaluate manipulation robustness under controlled simulation settings. These tasks involve variations in object instances, initial configurations, and scene layouts, requiring the agent to generalize across diverse visual and spatial conditions. Each task is defined with a fixed goal specification, while the initial states are randomized to assess policy robustness.
As the benchmark does not provide an official training set, following prior works~\cite{qu2025spatialvla, kwok2025robomonkey}, we train our model on the BridgeData V2 dataset~\cite{walke2023bridgedata, kwok2025robomonkey}, which contains large-scale real-world robot manipulation trajectories. The trained policy is then directly evaluated in the SimplerEnv simulator without additional fine-tuning, testing its cross-domain generalization ability from real-world data to simulation.
For evaluation, we adopt the standard success rate metric, where a trial is considered successful if the target object is accurately grasped and placed at the designated location within a fixed time horizon. Each task is evaluated over multiple randomized episodes to ensure statistical reliability. The simulation setup follows the default configuration of SimplerEnv, including camera viewpoints, control frequency, and action space, ensuring a fair comparison with prior methods~\cite{kwok2025robomonkey,nakamoto2024steering,jang2025verifier}.
All experimental protocols are consistent with prior works~\cite{kwok2025robomonkey,nakamoto2024steering,jang2025verifier} to ensure comparability.

\ptitle{LIBERO.}
The LIBERO benchmark~\cite{liu2023libero} comprises a diverse suite of long-horizon manipulation tasks designed to evaluate the generalization and compositional reasoning capabilities of robotic policies. It is organized into four subsets: \textit{LIBERO-Spatial}, \textit{LIBEROObject}, \textit{LIBEROGoal}, and \textit{LIBEROLong}, each targeting a different aspect of generalization. The Spatial subset focuses on variations in spatial arrangements, the Object subset evaluates generalization across different object instances, the Goal subset introduces variations in task objectives, and the Long subset contains multi-stage tasks requiring extended sequential reasoning and planning.
Each task in LIBERO is defined as a sequence of manipulation primitives, where the agent must complete multiple sub-goals in the correct order to achieve the final objective. The environments exhibit diversity in object layouts, appearances, and initial conditions, requiring the policy to generalize beyond memorized behaviors. In particular, the Long subset presents significant challenges due to compounding errors over extended horizons.
For training, we finetune our model on the official LIBERO training dataset~\cite{liu2023libero}, which provides demonstration trajectories for all task subsets. During the evaluation phase, the trained model is deployed in the simulation environment and its performance is judged based on the success rate of task completion. Following~\cite{jang2025verifier}, we reported the average success rate (\%) in 4 task suites. Each task suite consisted of 10 tasks, and each task was tested 50 times. The results of our method were obtained by averaging the results based on 3 random seeds.
To ensure fair comparison, we adopt the default simulation settings provided by the benchmark~\cite{liu2023libero}. All experiments are conducted under identical conditions across different task subsets.

\subsection{Additional Real-world Settings}

\begin{figure}[htbp]
    \centering{\includegraphics[width=1.0\linewidth]{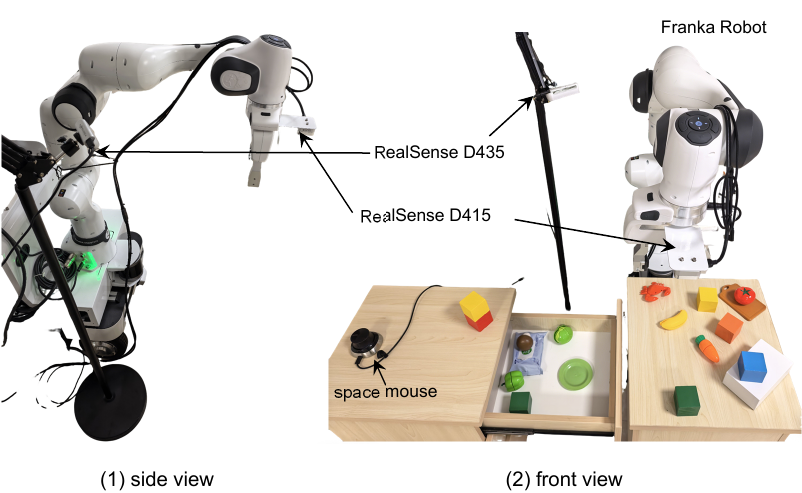}}  
    \caption{
    The scene setting of the real-world experiments.
    }
    \label{fig:robot}
\end{figure}

\ptitle{Hardware Setup.}
All real-world experiments are conducted on a Franka robot. 
The system is equipped with two RGB cameras for visual observation: a fixed third-person camera (Intel RealSense D435) providing a global view of the workspace, and a wrist-mounted camera (Intel RealSense D415) attached to the end-effector for close-range perception. 
The workspace consists of a tabletop setup with two cabinets containing drawers, along with multiple everyday objects placed in different regions. 
All components are synchronized and controlled via the Robot Operating System (ROS), enabling real-time perception and control. The front view and side view of the entire real-world experimental scene are shown in Figure~\ref{fig:robot}.

\ptitle{Task Design.}
We construct both in-distribution (ID) and out-of-distribution (OOD) manipulation tasks to evaluate the robustness and generalization capability of our method. 
The ID setting includes four pick-and-place tasks with variations in object positions and spatial layouts. 
These tasks share similar object categories and workspace configurations as the training data.
Furthermore, to evaluate the generalization ability of our method under varying spatial placements (e.g., height and position), and to demonstrate the effectiveness of its spatial reasoning capability, we place the plate at three different heights: on the tabletop, on top of a single cube, and on top of two stacked cubes.
In contrast, the OOD setting introduces previously unseen objects (e.g., banana) as well as novel spatial configurations (e.g., target object in the drawer). 
In total, the OOD benchmark consists of two tasks designed to evaluate zero-shot generalization.

\ptitle{Data Collection.}
For each in-distribution task, we collect 50 demonstration trajectories using teleoperation. 
Each trajectory contains approximately 500~700 timesteps, covering the full sequence from grasping to placement. 
The demonstrations capture diverse initial states and object arrangements to improve policy robustness. 
No additional training data is collected for OOD tasks.

\ptitle{Evaluation Protocol.}
During evaluation, each task is executed over 50 trials with randomized initial object configurations. 
A trial is considered successful if the robot correctly grasps the target object and places it at the designated location within a predefined time horizon.
For OOD tasks, we additionally vary object heights and spatial positions to assess the model's ability to generalize to unseen conditions. 
The evaluations of different methods are all conducted under the same experimental conditions.

\subsection{Comparison with 3D-VLAs.}
As summarized in Table~\ref{3DVLA}, VeriSpace consistently outperforms existing 3D-VLAs~\cite{li20253ds, qu2025spatialvla} on the SimplerEnv benchmark by achieving an average success rate of 55.0\%, which represents a 12.3\% absolute improvement over SpatialVLA. Notably, our method demonstrates exceptional robustness on spatially demanding tasks. On "Stack Cubes" and "Spoon on Towel", VeriSpace substantially surpasses the baselines by large margins with success rates of 62.0\% and 48.0\% respectively. Although SpatialVLA overfits to the simpler "Eggplant in Basket" task with a perfect score of 100.0\%, it struggles significantly with the remaining scenarios. Conversely, VeriSpace maintains a balanced and superior generalization capability across tasks with varying levels of spatial complexity.

\begin{table}[htbp]
    \centering
    \scriptsize
    \caption{Comparison with 3D-VLAs on SimplerEnv.}
    \label{3DVLA}
    \resizebox{1.0\linewidth}{!}{
    \begin{tabular}{l|cccc|c}
    \toprule
    Methods         &Eggplant in Basket & Carrot on Plate  &Stack Cubes  &Spoon on Towel & \textbf{Average}  \\
    \midrule
    3DS-VLA~\cite{li20253ds}   &74.0 &32.0 &28.0 &22.0 &39.0\\
SpatialVLA~\cite{qu2025spatialvla}   &\textbf{100.0} &25.0 &29.2 &16.7 & 42.7\\
    \rowcolor{linecolor2}{\textbf{VeriSpace (Ours)}}  & 76.0 & \textbf{34.0} & \textbf{62.0} & \textbf{48.0} & \textbf{55.0}\\
    \bottomrule
    \end{tabular}
    }
\end{table}

\subsection{Runtime Analysis}\label{Runtime}
We analyze the runtime performance of our framework on the SimplerEnv benchmark using OpenVLA as the VLA baseline. 
The runtime results are summarized in Table~\ref{tab:runtime}.
Although our framework involves repeated action sampling and verification, the overall computational overhead increases little, introducing only an additional 0.15 seconds with verification strategy. This efficiency is achieved through several optimizations. First, due to redundancy in repeated action sampling, we follow~\cite{kwok2025robomonkey} to reuse KV caches and implement the sampling pipeline using SGLang, decreasing latency from 1.59s to 0.47s. 
Second, for real-world deployment, our client-server deployment ensures real-time execution. Moreover, our inherently \textbf{parallelizable} sample-and-verify strategy \textbf{matches generic VLA speeds} via  multi-GPU processing, reducing the latency from 1.59s to \textbf{0.28s}. 
Overall, our work focuses on test-time scaling laws, yielding significant performance gains (\textcolor{red}{\textbf{+18.0/ 32\%}}) at a manageable computational cost (\textcolor{red}{\textbf{+0.07s}}).


\begin{table}[h]
\centering
\caption{Runtime analysis for each stage.}
\label{tab:runtime}
\resizebox{1.0\linewidth}{!}{
\begin{tabular}{lcccc}
\toprule
Methods &  Sampling (s) & Verification (s)& Total (s) &Success rate (\%)\\
\midrule
OpenVLA & 0.21 & - & 0.21 & 37.0\\
OpenVLA w repeated sampling & 1.59 & - & 1.59 & 37.0\\
\rowcolor{linecolor2}{VeriSpace on 1 GPU} & {0.47} & {0.15} & \textbf{0.62} & \textbf{55.0}\\
\rowcolor{linecolor2}{VeriSpace on 8 GPUs} & {0.23} & {0.05} & \textbf{0.28} & \textbf{55.0}\\
\bottomrule
\end{tabular}
}
\end{table}

\subsection{Analysis on the Source of our Gains.}
As detailed in Table~\ref{Additional_Ablation}, we systematically dissect the performance improvements to clarify the source of our gains. First, the table shows that our method (Row 3) already improves the Generic VLA baseline (Row 1) even without 3D input, yielding absolute increases of \textbf{\textcolor{red}{+5.0/+5.3}} on SimplerEnv and Real-world tasks, respectively. This confirms the inherent effectiveness of our proposed generate-and-verify paradigm in facilitating logical task execution. Second, when 3D representations are introduced (Row 4), the overall gain surges significantly to \textbf{\textcolor{red}{+18.0/+14.0}}. This substantial leap indicates that the improvement is not simply from adding another sensory modality, {but stems directly from our novel 3D-aware modules}, which are inherently designed for and strictly rely on 3D inputs. Specifically, critical components including the 3D-Injected Scene Encoding \textbf{(\textcolor{red}{+8.5})} and Action Reasoning \textbf{(\textcolor{red}{+6.5})} are the primary drivers of this spatial understanding, which is further corroborated by the detailed ablations in Table \ref{Ablation}. Finally, to isolate the impact of model capacity, we compare our full pipeline with RoboMonkey~\cite{kwok2025robomonkey}. Notably, both approaches introduce exactly 7B extra parameters. Despite sharing the identical parameter scale, VeriSpace significantly outperforms RoboMonkey by massive margins of \textbf{\textcolor{red}{+15.5/+12.9}}, providing definitive evidence that our gains are fundamentally driven by the architectural and algorithmic design rather than mere increases in model capacity.

\begin{table}[htbp]
    \centering
    \caption{Additional ablation study on the source of our gains.}
    \resizebox{1.0\linewidth}{!}{
    \begin{tabular}{l|l|cc|c}
    \toprule
    No.  &Methods        & SimplerEnv  & Real-world  & Extra parameters\\
    \midrule
    1)   & Generic VLA  &37.0 & 58.3  &-\\
    2)   &+ RoboMonkey  &39.5 & 59.4  &7B\\
    \rowcolor{linecolor2}{3)}   &+ VeriSpace w/o 3D input  &42.0 & 63.6  &7B\\
    \rowcolor{linecolor2}{4)}         &+ VeriSpace w 3D input   &\textbf{55.0}  &\textbf{72.3} &7B\\
    \bottomrule
    \end{tabular}
    }
    \label{Additional_Ablation}
\end{table}

\subsection{Qualitative Results}

\begin{figure*}[t]
    \centering{\includegraphics[width=1.0\textwidth]{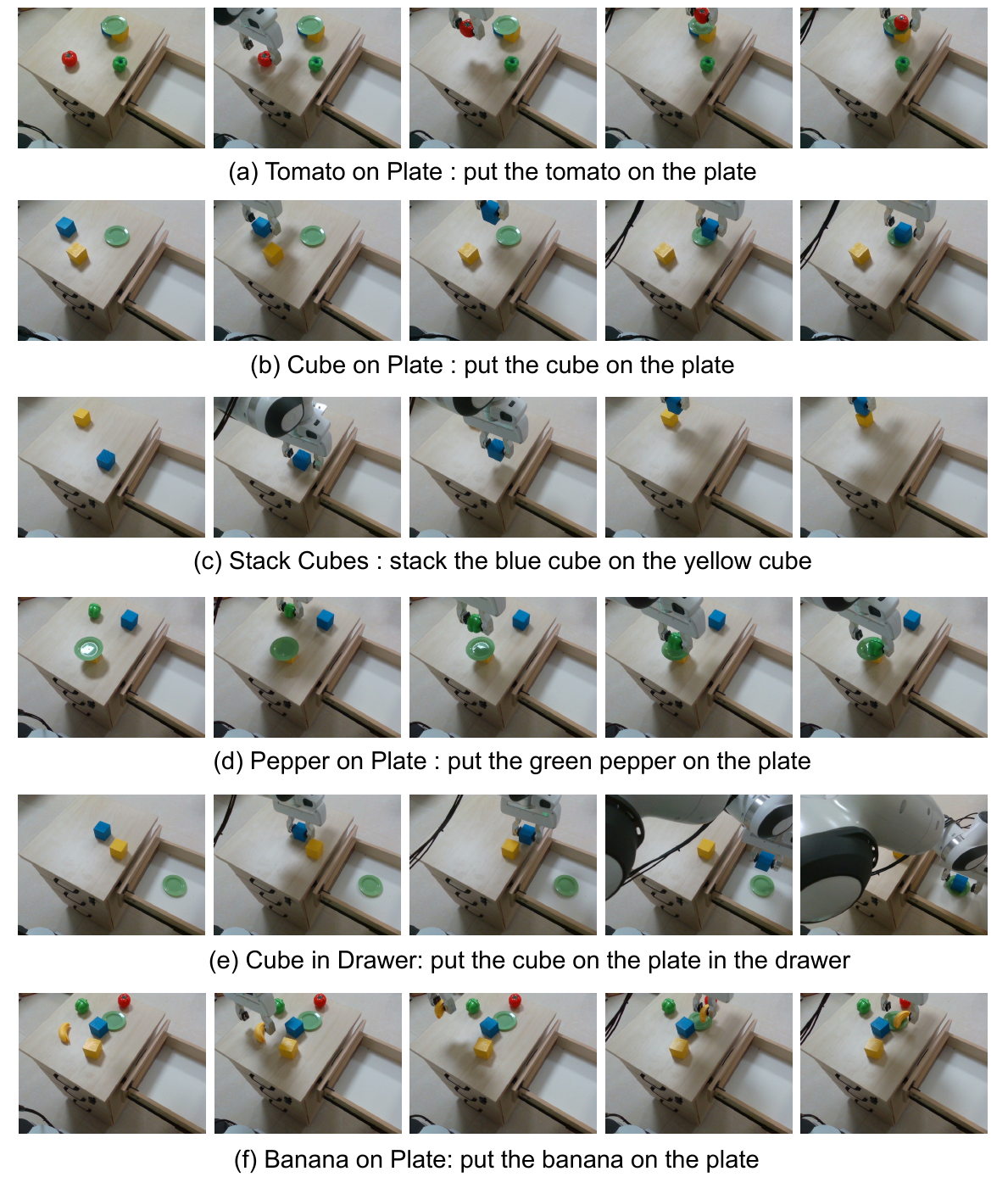}}  
    \caption{
    Qualitative results of our VeriSpace on real-world tasks. Representative examples include \textit{Tomato on Plate}, \textit{Cube on Plate}, \textit{Stack Cubes}, \textit{Pepper on Plate}, \textit{Cube in Drawer} and \textit{Banana on Plate} tasks.
    }
    \label{qr:realworld}
\end{figure*}

\ptitle{Qualitative Results in Real-world.}
We present qualitative results of VeriSpace on a diverse set of real-world manipulation tasks, as shown in Figure~\ref{qr:realworld}. Representative examples include \textit{Tomato on Plate}, \textit{Cube on Plate}, \textit{Stack Cubes}, \textit{Pepper on Plate}, \textit{Cube in Drawer}, and \textit{Banana on Plate}. In Figure~\ref{qr:realworld}, (a) to (d) represent the ID tasks, and (e) and (f) represent the OOD tasks. These tasks cover a range of challenges, including precise placement, object stacking, and varying spatial configurations such as drawers. The results demonstrate that VeriSpace can reliably infer spatial relationships and execute multi-step manipulation behaviors under varying spatial configurations. In particular, the model exhibits strong robustness to changes in object geometry, placement height, and scene layout, highlighting its ability to generalize across diverse real-world scenarios.

\begin{figure*}[t]
    \centering{\includegraphics[width=1.0\textwidth]{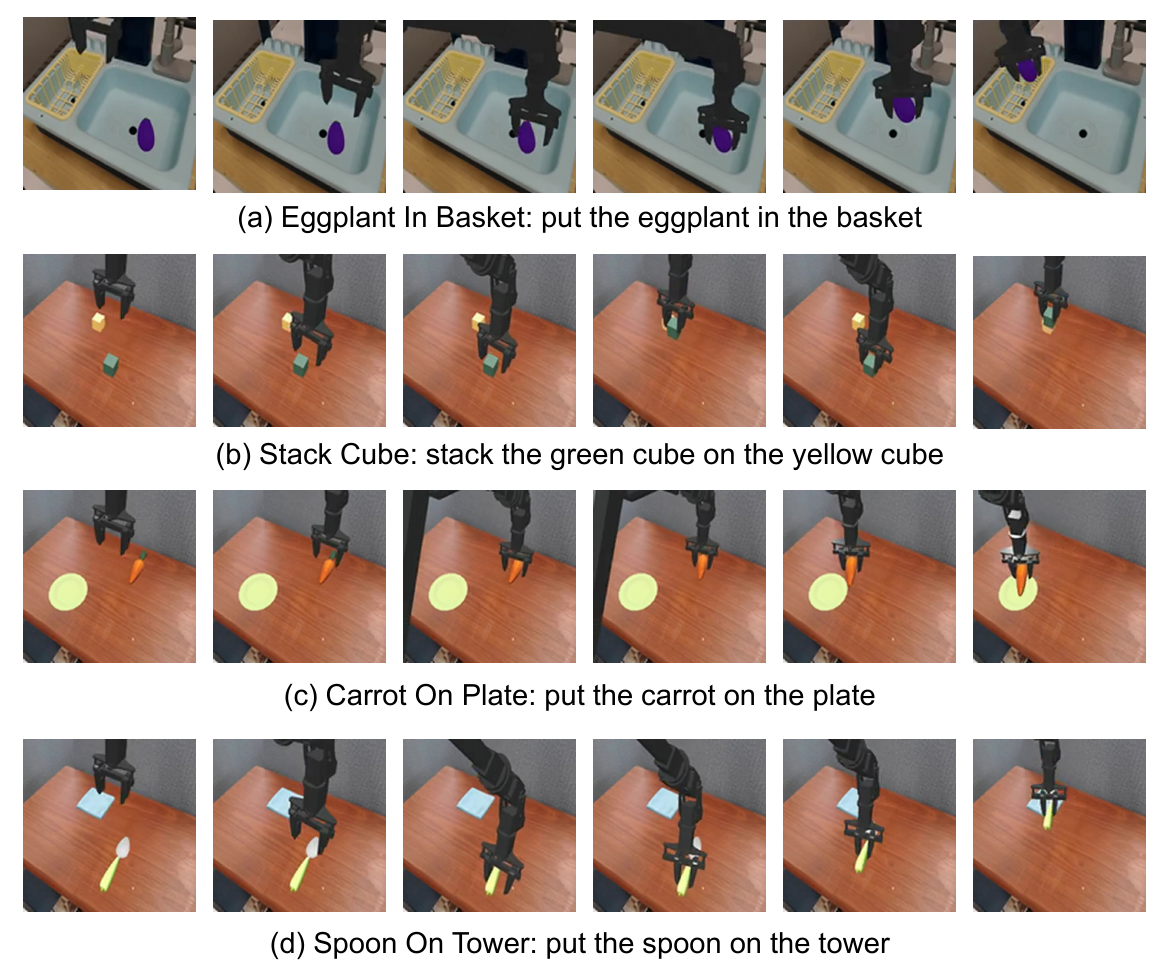}}  
    \caption{
    Qualitative results of our VeriSpace on the SimplerEnv-WidowX benchmark~\cite{li24simpler}. Representative examples include \textit{Eggplant In Basket}, \textit{Stack Cube}, \textit{Carrot On Plate} and \textit{Spoon On Tower} tasks.
    }
    \label{qr:simpler}
\end{figure*}

\ptitle{Qualitative Results in Simulation.}
We present qualitative results of VeriSpace on the SimplerEnv-WidowX benchmark~\cite{li24simpler}, as illustrated in Figure~\ref{qr:simpler}. Representative examples include \textit{Eggplant In Basket}, \textit{Stack Cube}, \textit{Carrot On Plate}, and \textit{Spoon On Tower}. These tasks involve diverse manipulation skills such as object placement, stacking, and handling varying spatial constraints. The results show that VeriSpace can effectively capture task-relevant spatial relationships and achieve accurate action scoring and candidate action selection. Notably, the model demonstrates robustness to variations in scene configuration, indicating its ability to generalize across different manipulation scenarios in simulation.

\section{Additional Related Works}

\subsection{3D-Enhanced VLA Models}
With the rapid development of large language models~\cite{touvron2023llama, brown2020language, ouyang2022training} and multimodal large language models~\cite{GPT4Vision23,llava23,DreamLLM23,ShapeLLM24}, and benefiting from the pretraining of vision–language models, Vision-Language-Action (VLA) models have gradually become one of the mainstream approaches for general robotic manipulation. 
Although these methods achieve impressive instruction-following capabilities, they are limited by 2D inputs and therefore lack 3D spatial perception and reasoning abilities. As a result, their performance in fine-grained spatial manipulation is restricted, and they are unable to perform 3D collision perception, leading to reduced robustness and safety. To address these issues, recent work has incorporated 3D perception into VLM fine-tuning and policy learning. For example, GeoVLA~\cite{sun2025geovla}, PointVLA~\cite{li2026pointvla}, 4D-VLA~\cite{zhang20254d}, and BridgeVLA~\cite{li2025bridgevla} directly use explicit RGB-D observations, introducing 3D information to enhance spatial perception. However, these approaches require additional 3D sensor configurations. 
Meanwhile, 3D-VLA~\cite{zhen20243d}, SpatialVLA~\cite{qu2025spatialvla}, and Evo-0~\cite{lin2025evo} extract 3D information from images using 3D foundation models and depth estimation models, and integrate 3D features into VLMs via a 3D encoder. However, such approaches may disrupt the pretrained visual–language alignment of VLMs. Our VeriSpace introduces 3D information at test time through sampling-verification method, thereby avoiding the aforementioned issues. Moreover, by leveraging a verifier with 3D spatial reasoning capabilities, our method also improves the robustness and safety of robotic manipulation.

\bibliographystyle{ACM-Reference-Format}
\bibliography{sample-base}

\end{document}